\def\year{2022}\relax
\documentclass[letterpaper]{article} % DO NOT CHANGE THIS
\usepackage{aaai22}  % DO NOT CHANGE THIS
\usepackage{times}  % DO NOT CHANGE THIS
\usepackage{helvet}  % DO NOT CHANGE THIS
\usepackage{courier}  % DO NOT CHANGE THIS
\usepackage[hyphens]{url}  % DO NOT CHANGE THIS
\usepackage{graphicx} % DO NOT CHANGE THIS
\usepackage{natbib}  % DO NOT CHANGE THIS AND DO NOT ADD ANY OPTIONS TO IT
\usepackage{caption} % DO NOT CHANGE THIS AND DO NOT ADD ANY OPTIONS TO IT
\DeclareCaptionStyle{ruled}{labelfont=normalfont,labelsep=colon,strut=off} % DO NOT CHANGE THIS
\usepackage{algorithm}
\usepackage{algorithmic}

\usepackage{newfloat}
\usepackage{listings}
\floatstyle{ruled}
\newfloat{listing}{tb}{lst}{}
\floatname{listing}{Listing}
\usepackage{amsmath}
\usepackage{enumitem}
\usepackage{array}
\usepackage{subcaption}
\usepackage{multirow}
\DeclareMathOperator*{\argmax}{argmax} % no space, limits underneath in displays

\title{An Online Data-Driven Emergency-Response Method for Autonomous Agents in Unforeseen Situations}
\author {
    Glenn Maguire,\textsuperscript{\rm 1}
    Nicholas Ketz, \textsuperscript{\rm 2}
    Praveen Pilly \textsuperscript{\rm 2}
    Jean-Baptiste Mouret \textsuperscript{\rm 1}
}
\affiliations {
    \textsuperscript{\rm 1} Inria, CNRS, Université de Lorraine\\
    \textsuperscript{\rm 2} Center for Human-Machine Collaboration, Information and Systems Sciences Laboratory, HRL Laboratories\\
    glenn.maguire@inria.fr, naketz@hrl.com, pkpilly@hrl.com, jean-baptiste.mouret@inria.fr
}
\begin{document}

\maketitle

\begin{abstract}

Reinforcement learning agents perform well when presented with inputs within the distribution of those encountered during training. However, they are unable to respond effectively when faced with novel, out-of-distribution events, until they have undergone additional training. This paper presents an online, data-driven, emergency-response method that aims to provide autonomous agents the ability to react to unexpected situations that are very different from those it has been trained or designed to address. In such situations, learned policies cannot be expected to perform appropriately since the observations obtained in these novel situations would fall outside the distribution of inputs that the agent has been optimized to handle. The proposed approach devises a customized response to the unforeseen situation sequentially, by selecting actions that minimize the rate of increase of the reconstruction error from a variational auto-encoder. This optimization is achieved online in a data-efficient manner (on the order of 30 data-points) using a modified Bayesian optimization procedure. We demonstrate the potential of this approach in a simulated 3D car driving scenario, in which the agent devises a response in under 2 seconds to avoid collisions with objects it has not seen during training.

\end{abstract}

\section{Introduction}

Recent advances in Reinforcement Learning (RL) through deep neural networks have shown promising results in developing autonomous agents that learn to effectively interact with their environments in a number of different application domains \citep{1, 2}, including learning to play video games from image pixels \citep{3, 4, 5}, generating optimal control policies for robots \citep{6, 8}, speech recognition and natural language processing \citep{9}, as well as making optimal trading decisions given dynamic market conditions \citep{7}. Under the RL paradigm, the agent learns to perform a given task through numerous training episodes involving trial-and-error interactions with its environment. By discovering the consequences of its actions in terms of the rewards obtained through these interactions the agent eventually learns the optimal policy for the given task.

These approaches work well in situations where it can be assumed that all the events encountered during deployment arise from the same distribution on which the agent has been trained. However, agents that must function within complex, real-world environments for an extended period of time can be subjected to unexpected circumstances outside of the distribution they have been designed for or trained on, due to environmental changes that arise. For example: an autonomous driving car may encounter significantly distorted lane-markings that it has never experienced before due to construction or wear, and must determine how to continue to drive safely; an unaware worker in a manufacturing facility may suddenly place a foreign object, such as their hand, within the workspace of a vision-guided robot-arm that must then react to avoid damage/injury; or the seam being tracked by a welding robot may suddenly shift due to a poorly secured work-piece, requiring the robot to react to minimize part-damage. In such unexpected, novel situations the agent's policy would be inapplicable, causing the agent to perhaps take unsafe actions.

In this paper, we consider scenarios where a trained agent encounters an unforeseen situation during deployment that renders available system or state-transition models highly unreliable, so that any inferences based on such models, as well as any pre-defined safe state/action regions, are no longer valid for safe decision-making. An agent unable to respond effectively to a novel situation when first encountered is vulnerable to take dangerous actions. This is of particular concern in safety-critical applications where sub-optimal actions can lead to damage or destruction of the agent or loss of human life.

While continual learning approaches exist \citep{29, 30, 31, 32, 33} that retrain agents, or re-optimize models, to eventually improve performance on novel events, they require a significant amount of time and repeated encounters with those events before learning the optimal policy that covers those cases. If these novel events represent dangerous situations in safety-critical applications, this still leaves the question of how to respond to these events in the interim.

We address this problem by developing a data-driven danger-mitigation system that allows an agent to deal with novel situations without reliance on the accuracy of existing models, or the validity of safe states and recovery policies developed offline or from past experiences. The key insight to our approach is that uncertainty in observations from the environment can be used not only to detect dangerous scenarios, but also as a driver for the generation of effective, short-term responses online, when necessary, to circumvent dangers, so that the agent can continue to function within its environment.

Increased observation uncertainty (e.g., as measured by out-of-bounds auto-encoder reconstruction errors \citep{16}) can be used to detect novelty, for which the existing policy is unprepared. It stands to reason, then, that decreasing this uncertainty would decrease novelty and return states to those that the current policy can handle effectively, thus correlating uncertainty-minimization with safety. For example, when an aerial vehicle descends too close to the ground and finds difficulty in maintaining control due to strong ground-effect, ascending back to milder conditions that it is better designed to handle returns it to safety. While use of uncertainty to detect potential danger is not new, using it to generate actions in an online manner, customized to the particular never-before-seen emergency as it unfolds, is novel.

In the absence of a reliable model or policy network to make proper action decisions, determination of an appropriate response to a novel situation must be data-driven and sequential. In other words, after an action is taken by the agent at a given time-step, the resulting measurements must be analyzed, along with all prior measurement data gathered so far during the course of the response, in order to determine the best action to take for the subsequent time-step. Moreover, in an emergency situation this response must be devised efficiently (i.e., in just a few time-steps), meaning that little data will typically be available for finding the optimal actions to take. This reactive approach, therefore, necessitates a fast, online, optimal decision-making method.

Bayesian Optimization (BO) provides an ideal theoretical framework for this type of problem scenario \citep{24}. BO is a data-efficient, global-optimization method for sequential decision-making where the objective function is expensive to evaluate or is taken to be a black-box. It uses a probabilistic approach that involves building and maintaining a distribution of functions to model the objective being optimized, and sequentially improving this probabilistic model through measurement data obtained online. This model is used to compute the next best action to take in a manner that balances exploration of the unknown regions of the objective and exploitation of regions found to be most likely to contain the optimal value.

Using this framework we devise an emergency response generation method that combines a modified BO procedure for efficient sequential optimization, with Gaussian Process (GP) regression for representing the probabilistic model of the objective. The objective function in our approach is a metric designed to capture the uncertainty in the observations obtained by the autonomous agent in a way that facilitates the generation of an effective emergency response. The responses generated by this method are intended to be action-sequences over a short time-span that are only initiated when deemed necessary to circumvent a dangerous situation that the agent is not yet prepared to handle. As such, our overall approach is referred to as the data-driven, emergency-response method.

\section{Preliminaries}

\subsection{Related Work}

While there are existing works related to safety for autonomous agents \citep{10, 11, 12, 13, 14, 15, 16, 17, 21}, they are unable to address the scenarios being considered in this paper, which are situations that are unexpected and outside of the distribution of events that the agent has been trained for. Typical safe decision-making strategies involve incorporating pre-designed penalties into the reward or cost function for actions deemed unsafe or dangerous when training a deep neural network to generate policies \citep{10, 11}, or restricting agent actions to ``safe" regions in order to prevent it from reaching unsafe states \citep{12, 13, 14, 15}. Such approaches are still vulnerable to unforeseen events not accounted for through the penalties or pre-determined safe regions stipulated.

Other approaches use examples of dangers in offline training in representative environments to either help identify potentially dangerous situations online and conservative behaviors to use based on pre-specified rules \citep{16}, or to learn recovery policies for specific dangerous scenarios that can be applied during deployment \citep{17}. During long-term deployment in complex environments, however, significantly novel events very different from those experienced during offline training can arise. These can produce potentially dangerous scenarios not effectively captured by the above-mentioned mechanisms, particularly when such situations develop very rapidly, thereby requiring a customized response to handle.

An agent must therefore be able to continually learn and adapt to such novel situations. Continual learning approaches in the literature that address this need, though, do so through the initiation of a new learning phase, with primary concerns being in trying to learn new tasks efficiently, in an autonomously triggered manner, while maintaining good performance on tasks already learned \citep{29, 30, 31, 32, 33}. Adaptation to the novel situation, then, is not instantaneous, and must happen over an extended period of time dictated by the continual learning method used.

Much work also exists in the literature on detection of novel situations, where observations outside of the distribution on which a deep learning agent has been trained must be identified \citep{16, 29, 33, 34}. This is typically done through establishing some measure of uncertainty in the output of the model or deep neural networks involved. In \citet{16}, for example, the average squared pixel error between the input and reconstructed observation image from an auto-encoder is compared to a threshold to detect a novel situation. Similarly, in \citet{29}, novelty detection is done through testing for a statistically significant difference in mean auto-encoder reconstruction error between what is expected on an already learned environment and what is sampled online. However, how to best respond to the novel event, particularly when it represents an emergency situation, is an open problem.

Nevertheless, what existing approaches do show is that deep learning neural networks produce erratic and unreliable predictions when presented with inputs very different from their training scenarios \citep{16, 18}, but also that uncertainty in predictions from such out-of-distribution inputs can be an effective way to detect novelty \citep{12, 16, 19}. Moreover, trying to jointly optimize for task performance and safety-violation can lead to restrictive, sub-optimal policies \citep{17, 20}. In addition, despite their limitations, these prior works also make it clear (as shown through the experiments on standard simulation benchmarks in \citet{14}, for example) that including a safety mechanism to assist learning agents that either limits or completely avoids dangerous actions improves success rate, constraint satisfaction, and sample efficiency.

\subsection{Summary of Bayesian Optimization}

Bayesian Optimization (BO) \citep{24} finds use as an efficient, global optimization technique for scenarios where the objective function, \( f(\textbf{x}) \), is significantly expensive to evaluate and/or is treated as a black-box system. BO builds a surrogate model of \( f(\textbf{x}) \) in the form of a probability distribution over this function that includes a mean function, \( \mu(\textbf{x}) \), representing the current best estimate of \( f(\textbf{x}) \) over the domain of $ f $, and a variance function, \( \sigma^2(\textbf{x}) \), representing the uncertainty in this estimate. 

This model is built sequentially, by sampling the input at each step that is most likely to produce a better value for $ f $ than the best found so far. With each data-point obtained, a revised posterior distribution over the function \( f(\textbf{x}) \) is computed. As such the model tends to improve only in the region that has the highest likelihood of containing the optimal solution. This makes BO data-efficient, allowing it to find a near-optimal solution with only a few evaluations.

In the proposed method, this surrogate model is a Gaussian Process (GP) regression model, \( G[f(\textbf{x})] \) \citep{21}. GP regression operates in function-space by defining a distribution over functions, thereby providing a formalism for computing a mean function, \( \mu(\textbf{x}) \), and a variance function, \( \sigma^2(\textbf{x}) \), for a given set of data-points. This regression model can then be used to infer the function value, \( y^* = \mu(x^*) \), for some unobserved input, $ x^* $. Details on how to solve a GP regression model to make inferences can be found in the Technical Appendix, Section A.

To select each subsequent data-point to sample, BO optimizes an acquisition function, \( \alpha( x, G[f(\textbf{x})] ) \). This heuristic function quantifies the utility of any given input, $ x $, in terms of its potential for optimizing \( f(\textbf{x}) \). It is designed to capture the trade-off between exploitation (i.e., sampling near the best solution found so far) and exploration (i.e., sampling in unexplored regions with greater uncertainty) \citep{25}.

Thus, at any given step, $ i $, of the sequential optimization process of BO, we will have available some data-points, \( \textbf{Px}_i = \begin{bmatrix} x_1 & x_2 & ... & x_n \end{bmatrix}_i \), \( \textbf{Py}_i = \begin{bmatrix} y_1 & y_2 & ... & y_n \end{bmatrix}_i \), and the objective is to find the next best point, $ x^*_i $, to sample in terms of its utility as expressed by the acquisition function. This produces an acquisition function optimization problem that can be expressed as:
\begin{equation}
    \textrm{Find:} \quad x^*_i = \argmax_{x_i \in D} \alpha( x_i, G[f(\textbf{x})]_i ),
\end{equation}
where:
\begin{equation}
    G[f(\textbf{x})]_i = P(f(\textbf{x}) | \textbf{Py}_i, \textbf{Px}_i, \textbf{x}).
\end{equation}
Here, $ D $ is the domain of \( f(\textbf{x}) \) (i.e., the input space), and \( P(f(\textbf{x}) | \textbf{Py}_i, \textbf{Px}_i, \textbf{x}) \) is the GP regression model of \( f(\textbf{x}) \) based on given data \( \textbf{Px}_i \) and \( \textbf{Py}_i \), which can be solved to obtain an estimate, \( \mu(x_i) \), and uncertainty, \( \sigma^2(x_i) \), of observation $ y_i $ at some unknown location $ x_i $.

\section{Methodology}

\subsection{Problem Description}

We consider the data-driven emergency-response method to serve as an independent module that monitors a trained and deployed agent as it performs a given task. Within this scenario there may be instances where the agent encounters a situation is has never seen before that presents a danger if not acted upon properly. The agent's existing policy is unable to determine an appropriate response without further training and any environment models become unreliable.

Whenever such an unforeseen event is encountered, the agent is considered to be in an emergency situation for which an emergency response is required to mitigate the danger. Two key sub-problems need to be addressed here: (1) Emergency Detection; and (2) Response Generation (Figure 1).

\begin{figure}[t]
    \centering
    \includegraphics[width=0.9\columnwidth]{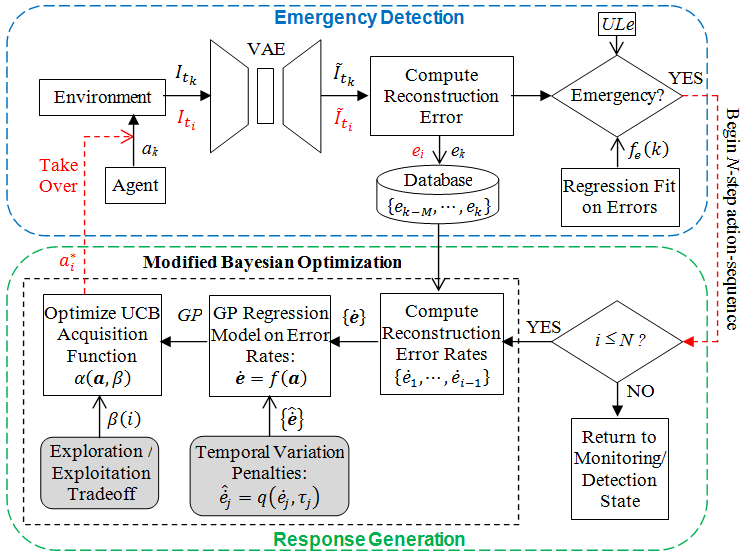}
    \caption{The data-driven emergency-response method.}
    \label{fig1}
\end{figure}

During deployment the observations from the environment as well as the actions that the policy intends to take are received by the Emergency Detection algorithm, which must decide whether or not an emergency situation is imminent. If it is not, then no further action is taken by the emergency-response system. Otherwise, the Response Generation algorithm is engaged, which takes over the policy's actions by replacing them with a customized action-sequence over the next $ N $ time-steps. This action-sequence must be generated online as the encounter with the novel situation unfolds.

\subsection{Emergency Detection}

When a novel situation arises, the corresponding observations will be very different from what the agent has seen during training or from past experiences. This increases the uncertainty associated with these novel observations. In our approach to emergency detection, we use a Variational Auto-Encoder (VAE) \citep{26} (see Technical Appendix, Section B for more details) to obtain a measure of observation uncertainty.

In particular, we represent observation uncertainty as the mean squared pixel-value error, $ e_t $, between the observation, $ I_t $, obtained from the environment at time $ t $, and the reconstructed output, \( \widetilde{I}_t \), from a VAE:
\begin{equation}
    e_t = \frac{1}{P} \sum_{k=1}^{P} (I_{t_k} - \widetilde{I}_{t_k})^2
\end{equation}
Here, \( I_{t_k} \) and \( \widetilde{I}_{t_k} \) are the \( k^\textrm{th} \) pixel intensity values of images $ I_t $ and \( \widetilde{I}_t \), respectively, and $ P $ is the total number of pixels in an image, including each of the R, G, and B color channels.

This VAE is assumed to be trained on the agent's past experiences or on a representative environment during pre-deployment under some nominal settings. This available experiential data can be used to empirically establish a threshold, \textit{ULe}, on the reconstruction errors observed during deployment, above which there is strong reason to believe that the agent has encountered a significantly novel situation.

This threshold is user-specified, and can be set to a reasonable value by observing the typical range of variation of errors under the nominal conditions in which the VAE has been trained. Qualitatively, a lower threshold corresponds to a more conservative stance on when an unforeseen situation is imminent, while a higher value would be more liberal, contributing to fewer false positives. An alternative probabilistic approach based on confidence limits is also given in the Technical Appendix, Section C.

The condition for detection is based on estimating the values of reconstruction errors into the short-term future using real-time data. At each time-step during deployment, a regression model, \( f_e(i) \), \( i \in [1,M] \), is fitted to the last $ M $ VAE reconstruction errors computed, which is then extrapolated $ K $ time-steps into the future. A positive detection occurs if at least the last $ q $ extrapolated errors exceed the established \textit{ULe} threshold (see Technical Appendix, Section C).

The choice for the values for $ q $ and $ K $ are related to how conservative or liberal one wishes to be when triggering the emergency response. The further into the future one extrapolates the more uncertain of the estimated values one would expect to be. To counteract this, a higher value for $ q $ can be chosen to provide a more stringent condition before concluding that a true positive detection event has occurred.

\subsection{Response Generation}

The proposed method for generating a response to address an unforeseen situation is based on the idea that taking actions that reduce the uncertainty in the observations should correspond to an effective response that guides the agent to a more familiar, and therefore relatively safer, state. In our approach, the response devised is an action-sequence that spans some fixed number of time-steps, $ N $. The generation of actions that reduce observation uncertainty is thus taken to be a sequential optimization process, where each action must ideally be the optimal decision to make given all the data gathered since the initiation of the response.

To perform this online optimization we use BO coupled with GP regression. As shown in Figure 1, the sequential optimization uses the rate of change of the VAE reconstruction errors to drive the BO loop at each time-step, $ i $, of the response action-sequence. This is because there may be situations where it may not be possible to find actions that reduce the reconstruction errors, and all that can be done is to minimize its increase. This would still be a valid response if that is the best that can be done given the circumstances. An imminent collision with an obstacle is again a good example – some situations may simply call for maximum braking as there may not be any way to swerve around the obstacle. In such cases, errors would only rise as the agent approached the obstacle, with braking helping to slow down the rate of increase until it eventually plateaus at a higher but stable value. Minimizing the error-rate would capture the need to slow down the rise of the errors in such situations, but would also be able to keep driving the errors down further (i.e., negative error-rates) if it is indeed possible.

The objective at each time-step, \( i \in [1,N] \), of the response is to find an action, \( a^*_i \), that minimizes the error-rate, \( \dot{e}_i \), that would result from that action, by conducting one cycle of the BO loop shown in Figure 1. This optimization will have available data-points containing all the actions, \( \textbf{Pa}_i = \begin{bmatrix} a_1 & a_2 & ... & a_{i-1} \end{bmatrix}_i \), taken in the last \( (i-1) \) time-steps of the action-sequence, as well as the corresponding true error-rates, \( \textbf{P}\dot{\textbf{e}}_i = \begin{bmatrix} \dot{e}_1 & \dot{e}_2 & ... & \dot{e}_{i-1} \end{bmatrix}_i \), that resulted.

The last $ M $ error data-points are always stored in a database. Once a response generation is triggered, every error data-point obtained from the start of the $ N $-step response is also saved (\( \textbf{Pa}_i \) and \( \textbf{P}\dot{\textbf{e}}_i \)) for the duration of the response. To compute the error-rate, \( \dot{e}_k \), corresponding to the \( k^\textrm{th} \) error data-point obtained, the available reconstruction errors, \( \textbf{e} \), in the data-set are first passed through a smoothing filter, $ f_s $, to compute the smoothed errors, \( \widetilde{\textbf{e}} \), as the raw data would be noisy. The last two smoothed error values can then be used to compute the rate, \( \dot{e}_k \), as:
\begin{equation}
    \dot{e}_k = \frac{f_s(k) - f_s(k-1)}{\delta i} = \widetilde{e}_k - \widetilde{e}_{k-1} .
\end{equation}

BO then proceeds to construct a model of the unknown relationship, \( \dot{\textbf{e}} = f(\textbf{a}) \), between error-rates, \( \dot{\textbf{e}} \), and actions, \( \textbf{a} \), for the given emergency scenario using GP regression. This GP model is used to conduct the relatively simpler acquisition function optimization to find the next best action, \( a^*_i \), to take, as described by equations (1) and (2).

The optimal action, \( a^*_i \), is then applied to the environment. At the subsequent time-step, the resulting error $ e_i $ will be obtained, from which \(\dot{e}_i\) can be computed. Both \( \textbf{Pa}_i \) and \( \textbf{P}\dot{\textbf{e}}_i \) are then updated accordingly and the above BO loop procedure is repeated, until the response length, $ N $, is reached.

\subsubsection{Acquisition Function:}

Several options for the acquisition function can be found in the literature (e.g., Expected Improvement, Entropy Search, and Knowledge Gradient). We employ the Upper Confidence Bound (UCB), given by Eq. (5). Here, \( \mu(\textbf{a}) \) and \( \sigma^2(\textbf{a}) \) are the mean and variance of the regression model for the relationship, \( \dot{\textbf{e}} = f(\textbf{a}) \).
\begin{equation}
    UCB = \alpha(\textbf{a}, \beta) = \mu(\textbf{a}) + \sqrt{\beta \cdot \sigma^2(\textbf{a})}.
\end{equation}
UCB is chosen since it includes a parameter, $ \beta $, that allows direct control over the balance between exploration and exploitation, that is, how much the system should try actions that are far from those already sampled versus how much should it focus on the most promising actions found so far.

Since an emergency response is time-critical, it is important to ensure a transition from an initial exploratory behavior to an exploitative one in a timely manner so that the search converges on an effective solution fast enough to avoid the danger. To accomplish this, the explicit parameter, $ \beta $, is set to a decreasing function of time, \( \beta(t_i) \), \(i \in [1,N] \). The initial value, $ \beta_0 $, must be relatively high to encourage the BO to explore the action-space. As the action-sequence progresses, this parameter should decrease to a relatively lower value, $ \beta_k $, so that the optimization begins to exploit the best solution found so far. These requirements produce the following constraints on the form of the time-varying function chosen for \( \beta(t_i) \):
\begin{gather}
    \beta (t_1) = \beta_0,\\
    \beta (t_i \geq t_k) = \beta_k, 1 < k \leq N,\\
    \beta_0 > \beta_k,\\
    \frac{d\beta(t_i)}{dt_i} \leq 0, \quad \forall t, \quad t_1 \leq t \leq t_N.
\end{gather}

In this way, the degree to which the BO explores initially can be controlled by the choice for $ \beta_0 $, and the degree to which it exploits the best solution found so far can be controlled through the choice for $ \beta_k $. The speed with which the optimization transitions from this exploration to exploitation behavior can be controlled by the choice for $ k $.

A second point of concern in devising the acquisition function is incorporating the influence of time. The underlying relationship between error-rate and actions can, in general, be expected to change with time. Thus, recent observations will have greater relevance to, and influence on, the decision being made at any given time-step compared to older observations. To account for this temporal variation, we propose a penalty function that discounts the utility of any given observation, as reflected by its corresponding acquisition function value, based on that observation's ``age" within the time-span of the response action-sequence.

However, the utility of any given action as determined by the UCB acquisition function depends on the GP regression model used to obtain \( \mu(\textbf{a}) \) and \( \sigma^2(\textbf{a}) \) (see Eq. (5)). The GP regression model captures the influence of past observations on any other unseen one being estimated based on their relative distances in action-space. Thus, the discounting of action utility must be incorporated into the error-rate data used to compute the GP regression model. As such, we define a penalty function that operates directly on the set of error-rates available at any given time-step of the response. In particular, at the \( i^{\textrm{th}} \) time-step of a response action-sequence, we will have available past observations given by \( \textbf{Pa}_i = \begin{bmatrix} a_1 & a_2 & ... & a_{i-1} \end{bmatrix}_i \) and \( \textbf{P}\dot{\textbf{e}}_i = \begin{bmatrix} \dot{e}_1 & \dot{e}_2 & ... & \dot{e}_{i-1} \end{bmatrix}_i \). Each error-rate, \( \dot{e}_j \), in \( \textbf{P}\dot{\textbf{e}}_i \) is transformed to a discounted measure, \( \hat{\dot{e}}_j \), through a penalty function, \( q(\dot{e}_j, \tau_j) \), before computing the GP regression, where:
\begin{align}
    \tau_j = i - j, \quad &\forall j, \quad 1 \leq j \leq (i-1), \textrm{and}\\
    \frac{dq}{d\tau} \geq 0, \quad &\forall \tau \geq 1.
\end{align}
Here, \( \tau_j \) represents the age of the \( j^{\textrm{th}} \) error-rate at time-step $ i $, and Eq. (11) indicates that the penalties should increase with age. This user-specified penalty function can be devised under this constraint depending on how strongly and quickly one wishes past data to lose its significance. An example is provided in the experiments presented in Section 4. 

Optimization of the acquisition function at each time-step can be conducted through standard approaches used in the literature. Typically, the UCB function can be effectively optimized using quasi-Newton methods such as the popularly-used L-BFGS-B algorithm \citep{23}.

\section{Simulation Experiments} \label{section4}

To demonstrate and validate the proposed method, experiments were conducted using the open-source CARLA autonomous driving car simulator \citep{27}. In these experiments, the proposed data-driven emergency-response method was used to detect and avoid imminent collisions with obstacles that an agent has never encountered before.

\subsection{Experimental Setup}

Nine different collision scenarios were setup in CARLA within a simulated urban environment (see Figure 2), each involving a different, unforeseen, stationary obstacle placed in the path of the autonomous car driving along a section of road in one of 5 different parts of the urban environment. These scenarios simulate a situation where an autonomous driving agent, assumed to have been trained to drive according to the rules of the road in an obstacle-free urban environment, is suddenly presented with an unforeseen situation involving a stationary obstacle placed in its path.

Normally the agent would not know how to best respond to such a novel situation and would have to experience it many times before learning the optimal response. This may even have to be repeated for different obstacle types. During this learning process the agent may take dangerous actions, possibly resulting in collisions with the obstacles.

\begin{figure}[t]
    \centering
    \includegraphics[width=0.9\columnwidth]{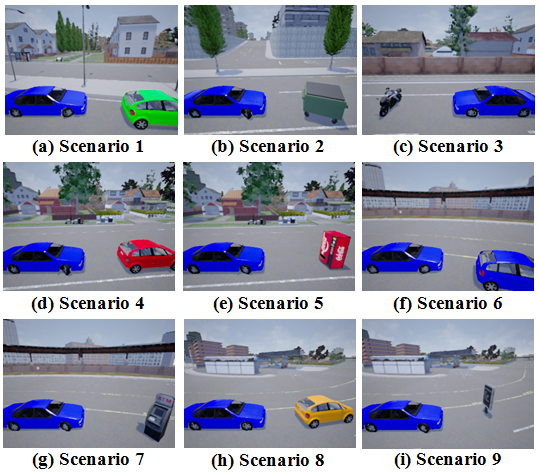}
    \caption{Simulated stationary-obstacle scenarios used in the experiments: (a) location A, green car; (b) location B, garbage container; (c) location B, motorcycle; (d) location C, red car; (e) location C, vending machine; (f) location D, blue car; (g) location D, ATM machine; (h) location E, orange car; (i) location E, street-sign.}
    \label{fig2}
\end{figure}

The proposed emergency-response method is incorporated into the simulation as an independent module that monitors the actions of the agent and the observations received from its forward-facing RGB camera sensor. Upon detecting an imminent collision using the approach described in Section 3.2 as the agent approaches a stationary obstacle, the module takes over the agent’s actions for the pre-specified next $ N $=30 time-steps with a customized action-sequence generated by the method described in Section 3.3 to attempt to prevent this collision.

At the start of each simulation frame, the agent receives an image corresponding to the current system state. The agent selects and applies an action, namely, the throttle, brake, and steering inputs, and the simulation updates the system state accordingly to the next time-step. This process repeats until the end of the simulation run. A worst-case scenario is simulated here, where, in the absence of the emergency-response system, the agent takes no action to avoid the obstacle and continues to follow the road. More details on the simulation setup can be found in the Technical Appendix, Section D.

A VAE is trained in a pre-deployment phase in this same environment without obstacles. The upper error threshold, \textit{ULe}, is empirically determined for each scenario based on observing the nominal reconstruction error variations obtained during the pre-deployment phase and selecting a reasonable value from this data.

At each time-step during deployment, a second order polynomial is fit to the last $ M $=15 reconstruction error data-points, and is extrapolated $ K $=7 time-steps into the future. A positive detection of an emergency situation occurs if all of the last $ q $=3 or more data-points exceed \textit{ULe}.

For the purposes of demonstration, the time-varying $ \beta $ parameter and the temporal variation penalties for the modified UCB acquisition function are given by equations (12) and (13), respectively, with $ \beta_0 $ = 0.07, $ \beta_k $ = 0, and $ k $ = 5. These satisfy the conditions given by equations (6) – (11), and allow the BO loop to make a rapid transition from exploration to exploitation in a manner sufficient to narrow down to an effective response in all the scenarios tested.
\begin{gather}
    \beta(ti) = -0.0028 t_i^2 + 0.07, \quad i \in [0, N-1],\\
    \begin{split}
        q(\dot{e}_j, \tau_j) = 6.15 ln(\tau_j) + 0.1, \\i \in [1, N-1], j \in [0, i-1].
    \end{split}
\end{gather}
It should be noted that while these functions and their parameters were chosen simply to demonstrate the overall method, they can always be further tuned and also learned online after initially starting with conservative values.

\subsection{Results}

In the first set of experiments, both the Emergency Detection and Response Generation components of the system were active and the average agent speed before encountering an obstacle was 20km/h. The proposed method (abbreviated as BO+GP) was compared to an alternate approach where the action-sequence was generated through a random selection of the action vector values at each time-step of the response. Multiple repetitions for each scenario were conducted (details in Technical Appendix, Section D) and the percentage of runs that resulted in successful collision avoidance (i.e., success-rate) was computed. Table 1 summarizes the results.

\begin{table}[t]
    \fontsize{9}{12}\selectfont
    \centering
    \begin{tabular}{|m{0.08\columnwidth}<{\centering}|m{0.15\columnwidth}<{\centering}|m{0.15\columnwidth}<{\centering}|m{0.15\columnwidth}<{\centering}|m{0.15\columnwidth}<{\centering}|}
        \hline
        \multirow{2}{0.08\columnwidth}{\textbf{Scen. \#}} & \multicolumn{2}{|c|}{\textbf{20km/h Tests}} & \multicolumn{2}{|c|}{\textbf{30km/h Tests}} \\
        \cline{2-5}
        & \textbf{BO+GP} & \textbf{Random} & \textbf{BO+GP} & \textbf{Random} \\
        \hline
        1 & 80\% & 10\% & 82\% & 5\% \\
        \hline
        2 & 65\% & 20\% & 90\% & 25\% \\
        \hline
        3 & 75\% & 35\% & 75\% & 10\% \\
        \hline
        4 & 75\% & 50\% & 80\% & 25\% \\
        \hline
        5 & 75\% & 25\% & 85\% & 70\% \\
        \hline
        6 & 45\% & 5\% & 55\% & 40\% \\
        \hline
        7 & 75\% & 30\% & 70\% & 25\% \\
        \hline
        8 & 70\% & 25\% & 25\% & 7.5\% \\
        \hline
        9 & 80\% & 15\% & 57.5\% & 17.5\% \\
        \hline
        \textbf{\textit{Avg.:}} & \textbf{71\%} & \textbf{24\%} & \textbf{68.8\%} & \textbf{25\%} \\
        \hline
    \end{tabular}
    \caption{Summary of success-rates for simulated collision-avoidance scenarios (note: taking no action resulted in a 0\% success-rate in all scenarios for both tests).}
    \label{table1}
\end{table}

To further elucidate the impact that the use of the proposed approach has on VAE reconstruction errors and error-rates, the experiments were repeated at a higher average agent speed of 30km/h and with only the Response Generation component of the proposed method active. For a fair comparison, both the proposed approach and the random-selection approach were manually triggered at the same time for all scenarios to ensure that the same distance and initial approach speed existed for both. Table 1 summarizes the success-rate results for these tests as well.

As a representative example for illustration, Figure 3 shows the plots of the variations in VAE reconstruction errors and error-rates over the span of the response action-sequences for scenario 1. The proposed approach is compared with both a random response and taking no action.
\begin{figure}[t]
    \centering
    \includegraphics[width=0.9\columnwidth]{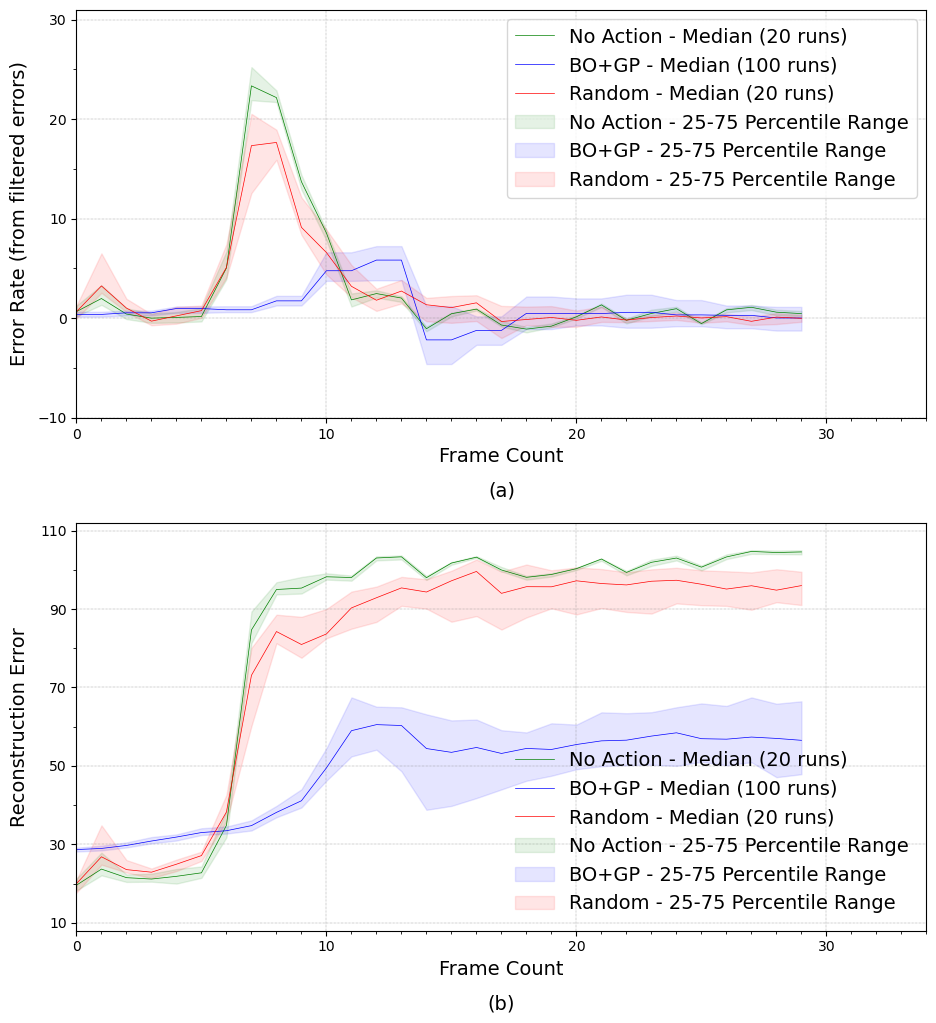}
    \caption{Comparison of the impact of the proposed approach (BO+GP), a random response, and no-action, on (a) VAE reconstruction error-rates (computed from filtered errors), and (b) VAE reconstruction errors, during the span of the response action-sequence initiated over the scenario 1 runs. Solid lines show the median value and the shaded region shows the \( 25^\textrm{th} \) percentile to \( 75^\textrm{th} \) percentile range of the plotted quantities among all the runs.}
    \label{fig3}
\end{figure}

\section{Conclusions and Discussion}

This paper proposed a data-driven emergency-response method for safe decision-making that allows a trained autonomous agent to safely address unforeseen situations encountered during deployment for which the existing policy becomes unreliable. When triggered, the method generates a response by finding optimal actions sequentially through minimization of VAE reconstruction errors from the novel observations using a modified BO algorithm. Simulation experiments in an autonomous car-driving domain demonstrate how minimization of observation uncertainty can find safe actions by using the proposed method to avoid collisions with obstacles the agent has never seen before.

The significantly greater average success-rate in collision-avoidance due to the proposed emergency-response method compared to a randomized approach, for both the average agent speeds tested, indicate that effective, intelligent actions are indeed being selected to avoid the novel dangerous situations, beyond simply what random chance would allow. This demonstrates how minimizing a measure of uncertainty in the observations can be correlated with good actions that help to effectively deal with unforeseen situations.

Figure 3(a) shows how the BO-based sequential optimization approach is able to quickly reduce and stabilize the error-rates over the course of the response. Through a combination of braking and turning, the agent not only reduces the speed with which it approaches the obstacle, thus significantly delaying the peak rate that results when the agent is at its closest proximity to the obstacle, but also maintains the error-rates at a much lower value throughout, compared to a random or no response. In fact, in some cases negative rates are also reached when the agent turns towards an adjacent lane, thereby side-stepping the obstacle and removing it from its view, causing the uncertainty measure to drop significantly. These effective, danger-avoiding behaviors are also reflected in the reconstruction errors themselves shown in Figure 3(b), where the errors rise at a relatively slower rate and plateau at a relatively lower final value due to the agent having transitioned to a more familiar (i.e., safer) state.

It is also evident that some scenarios presented more of a challenge than others. Detection of the imminent collisions was observed to happen when the agent was on average about 8m away from the obstacle. This left little distance and time to react, and in some cases it was not enough for the final response to avoid the collision, even though it may have been effective had the danger been detected sooner. Qualitative observations of some of the failures under the proposed method show that the agent still tries to make sensible maneuvers to avoid the collision and almost succeeds.

Curved roads are especially challenging, where a much stronger steering response may be needed to completely avoid the collision depending on which way the road turns and on which side the optimization decides to swerve (e.g., turning off on the side in the direction of the tangent to the road’s curvature would require less sharper steering to avoid a collision than choosing to turn into the curve). This accounts for the lower success-rates observed for the proposed method under some of the curved-road scenarios (scenarios 6-9). However, when the obstacle itself is smaller, the success-rate improves since the scenario is more forgiving of actions that may not have been as strong as one would ideally want them to be (e.g., compare the higher success-rates for the smaller ATM (scenario 7) and street-sign (scenario 9) obstacles to their wider and larger car-obstacle counterparts (scenarios 6 and 8, respectively) for the 30km/h tests).

It should be noted that the intention here was not to create the best obstacle-avoidance system for an autonomous car, but rather to demonstrate how minimization of observation uncertainty can be an effective driver to safely address novel situations that a learning agent would otherwise be unprepared to respond to. Moreover, the proposed method does not rely on having an accurate model of the environment or prior experience with those dangerous scenarios.

The method is also generic as it does not use context-specific information to generate the response (i.e., it does not need to know what the obstacle is, or what its presence means in the context of an autonomous driving car). As such, the performance of the method can always be improved by incorporating context-specific mechanisms on top of the basic emergency-response system for the particular application being addressed, if so desired.

\nocite{*}
\bibliography{references.bib}

\begin{thebibliography}{36}
\providecommand{\natexlab}[1]{#1}

\bibitem[{Achiam and Amodei(2019)}]{20}
Achiam, J.; and Amodei, D. 2019.
\newblock Benchmarking Safe Exploration in Deep Reinforcement Learning.
\newblock \url{https://d4mucfpksywv.cloudfront.net/safexp-short.pdf}.
\newblock In NeurIPS Deep Reinforcement Learning Workshop.

\bibitem[{Achiam et~al.(2017)Achiam, Held, Tamar, and Abbeel}]{11}
Achiam, J.; Held, D.; Tamar, A.; and Abbeel, P. 2017.
\newblock {Constrained Policy Optimization}.
\newblock In \emph{Proceedings of the 34th International Conference on Machine
  Learning - Volume 70}, 22–31. Cambridge, MA, USA: {JMLR.org}.

\bibitem[{Arulkumaran et~al.(2017)Arulkumaran, Deisenroth, Brundage, and
  Bharath}]{1}
Arulkumaran, K.; Deisenroth, M.~P.; Brundage, M.; and Bharath, A.~A. 2017.
\newblock {Deep Reinforcement Learning: A Brief Survey}.
\newblock \emph{IEEE Signal Processing Magazine}, 34(6): 26--38.

\bibitem[{Bengio et~al.(2015)Bengio, Vinyals, Jaitly, and Shazeer}]{9}
Bengio, S.; Vinyals, O.; Jaitly, N.; and Shazeer, N. 2015.
\newblock {Scheduled Sampling for Sequence Prediction with Recurrent Neural
  Networks}.
\newblock In \emph{Proceedings of the 28th International Conference on Neural
  Information Processing Systems - Volume 1}, 1171–1179. Cambridge, MA, USA:
  {MIT Press}.

\bibitem[{Brochu, Cora, and de~Freitas(2010)}]{25}
Brochu, E.; Cora, M.; and de~Freitas, N. 2010.
\newblock {A Tutorial on Bayesian Optimization of Expensive Cost Functions,
  with Application to Active User Modeling and Hierarchical Reinforcement
  Learning}.
\newblock Technical Report TR-2009-023, Department of Computer Science,
  University of British Columbia.

\bibitem[{Brown and Sandholm(2017)}]{5}
Brown, N.; and Sandholm, T. 2017.
\newblock {Libratus: The Superhuman AI for No-Limit Poker}.
\newblock In \emph{Proceedings of the Twenty-Sixth International Joint
  Conference on Artificial Intelligence {(IJCAI-17)}}, 5226--5228. Menlo Park,
  Calif: {IJCAI Organization}.

\bibitem[{Burr(1942)}]{35}
Burr, I.~W. 1942.
\newblock {Cumulative Frequency Functions}.
\newblock \emph{The Annals of Mathematical Statistics}, 13(2): 215--232.

\bibitem[{Caselles-Dupr{\'e}, Garcia-Ortiz, and Filliat(2021)}]{29}
Caselles-Dupr{\'e}, H.; Garcia-Ortiz, M.; and Filliat, D. 2021.
\newblock S-TRIGGER: Continual State Representation Learning via Self-Triggered
  Generative Replay.
\newblock International Joint Conference on Neural Networks (IJCNN 2021).
\newblock Accepted.

\bibitem[{Deng et~al.(2017)Deng, Bao, Kong, Ren, and Dai}]{7}
Deng, Y.; Bao, F.; Kong, Y.; Ren, Z.; and Dai, Q. 2017.
\newblock {Deep Direct Reinforcement Learning for Financial Signal
  Representation and Trading}.
\newblock \emph{IEEE Transactions on Neural Networks and Learning Systems},
  28(3): 653--664.

\bibitem[{Dosovitskiy et~al.(2017)Dosovitskiy, Ros, Codevilla, Lopez, and
  Koltun}]{27}
Dosovitskiy, A.; Ros, G.; Codevilla, F.; Lopez, A.; and Koltun, V. 2017.
\newblock {{CARLA}: {An} Open Urban Driving Simulator}.
\newblock In \emph{Proceedings of the 1st Annual Conference on Robot Learning},
  3521–3526. Bletchley Park, UK: {PMLR}.

\bibitem[{Eysenbach et~al.(2018)Eysenbach, Gu, Ibarz, and Levine}]{13}
Eysenbach, B.; Gu, S.; Ibarz, J.; and Levine, S. 2018.
\newblock {Leave no Trace: Learning to Reset for Safe and Autonomous
  Reinforcement Learning}.
\newblock In \emph{6th International Conference on Learning Representations,
  ICLR 2018, Vancouver, BC, Canada, April 30 - May 3, 2018, Conference Track
  Proceedings}, 1--18. La Jolla, CA, USA: {iclr.cc}.

\bibitem[{Fisac et~al.(2019)Fisac, Akametalu, Zeilinger, Kaynama, Gillula, and
  Tomlin}]{12}
Fisac, J.~F.; Akametalu, A.~K.; Zeilinger, M.~N.; Kaynama, S.; Gillula, J.; and
  Tomlin, C.~J. 2019.
\newblock {A General Safety Framework for Learning-Based Control in Uncertain
  Robotic Systems}.
\newblock \emph{IEEE Transactions on Automatic Control}, 64(7): 2737--2752.

\bibitem[{Francois-Lavet et~al.(2018)Francois-Lavet, Henderson, Islam,
  Bellemare, and Pineau}]{2}
Francois-Lavet, V.; Henderson, P.; Islam, R.; Bellemare, M.; and Pineau, J.
  2018.
\newblock {An Introduction to Deep Reinforcement Learning}.
\newblock \emph{IEEE Signal Processing Magazine}, 11(3-4): 219–354.

\bibitem[{Ha and Schmidhuber(2018)}]{36}
Ha, D.; and Schmidhuber, J. 2018.
\newblock World Models.
\newblock arXiv:1803.10122.

\bibitem[{Johnson, Kotz, and Balakrishnan(1994)}]{22}
Johnson, N.~L.; Kotz, S.; and Balakrishnan, N., eds. 1994.
\newblock \emph{Continuous Univariate Distributions. Vol. 2}.
\newblock Hoboken, NJ, USA: Wiley-Interscience.

\bibitem[{Kingma and Welling(2014)}]{26}
Kingma, D.~P.; and Welling, M. 2014.
\newblock Auto-Encoding Variational Bayes.
\newblock arXiv:1312.6114.

\bibitem[{Kirkpatrick et~al.(2017)Kirkpatrick, Pascanu, Rabinowitz, Veness,
  Desjardins, Rusu, Milan, Quan, Ramalho, Grabska-Barwinska, Hassabis, Clopath,
  Kumaran, and Hadsell}]{28}
Kirkpatrick, J.; Pascanu, R.; Rabinowitz, N.; Veness, J.; Desjardins, G.; Rusu,
  A.~A.; Milan, K.; Quan, J.; Ramalho, T.; Grabska-Barwinska, A.; Hassabis, D.;
  Clopath, C.; Kumaran, D.; and Hadsell, R. 2017.
\newblock {Overcoming catastrophic forgetting in neural networks}.
\newblock \emph{Proceedings of the National Academy of Sciences}, 114(13):
  3521--3526.

\bibitem[{Lee et~al.(2017)Lee, Kim, Jun, Ha, and Zhang}]{31}
Lee, S.-W.; Kim, J.-H.; Jun, J.; Ha, J.-W.; and Zhang, B.-T. 2017.
\newblock {Overcoming Catastrophic Forgetting by Incremental Moment Matching}.
\newblock In \emph{Advances in Neural Information Processing Systems},
  4652--4662. Red Hook, NY, USA: {Curran Associates, Inc.}

\bibitem[{Li, Kalabić, and Chu(2018)}]{15}
Li, Z.; Kalabić, U.; and Chu, T. 2018.
\newblock {Safe Reinforcement Learning: Learning with Supervision Using a
  Constraint-Admissible Set}.
\newblock In \emph{2018 Annual American Control Conference (ACC)}, 6390--6395.
  Piscataway, New Jersey, USA: {IEEE Press}.

\bibitem[{Liu and Nocedal(1989)}]{23}
Liu, D.; and Nocedal, J. 1989.
\newblock {On the limited memory BFGS method for large scale optimization}.
\newblock \emph{Mathematical Programming}, 45(1): 503–528.

\bibitem[{Manevitz and Yousef(2007)}]{19}
Manevitz, L.; and Yousef, M. 2007.
\newblock {One-class document classification via Neural Networks}.
\newblock \emph{Neurocomputing}, 70(7): 1466--1481.

\bibitem[{Marchi et~al.(2015)Marchi, Vesperini, Eyben, Squartini, and
  Schuller}]{33}
Marchi, E.; Vesperini, F.; Eyben, F.; Squartini, S.; and Schuller, B. 2015.
\newblock {A novel approach for automatic acoustic novelty detection using a
  denoising autoencoder with bidirectional LSTM neural networks}.
\newblock In \emph{2015 IEEE International Conference on Acoustics, Speech and
  Signal Processing (ICASSP)}, 1996--2000. Piscataway, New Jersey, USA: {IEEE
  Press}.

\bibitem[{Mnih et~al.(2015)Mnih, Kavukcuoglu, Silver, Rusu, Veness, Bellemare,
  Graves, Riedmiller, Fidjeland, Ostrovski, Petersen, Beattie, Sadik,
  Antonoglou, King, Kumaran, Wierstra, Legg, and Hassabis}]{3}
Mnih, V.; Kavukcuoglu, K.; Silver, D.; Rusu, A.~A.; Veness, J.; Bellemare,
  M.~G.; Graves, A.; Riedmiller, M.; Fidjeland, A.~K.; Ostrovski, G.; Petersen,
  S.; Beattie, C.; Sadik, A.; Antonoglou, I.; King, H.; Kumaran, D.; Wierstra,
  D.; Legg, S.; and Hassabis, D. 2015.
\newblock {Human-level control through deep reinforcement learning}.
\newblock \emph{Nature}, 518(7540): 529--533.

\bibitem[{Mockus(2013)}]{24}
Mockus, J., ed. 2013.
\newblock \emph{Bayesian Approach to Global Optimization: Theory and
  Applications}.
\newblock Boston, MA, USA: Kluwer Academic Publishers.

\bibitem[{Nguyen et~al.(2018)Nguyen, Li, Bui, and Turner}]{30}
Nguyen, C.~V.; Li, Y.; Bui, T.~D.; and Turner, R.~E. 2018.
\newblock {Variational Continual Learning}.
\newblock In \emph{Sixth International Conference on Learning Representations},
  1--18. La Jolla, CA, USA: {iclr.cc}.

\bibitem[{Pan et~al.(2017)Pan, You, Wang, and Lu}]{6}
Pan, X.; You, Y.; Wang, Z.; and Lu, C. 2017.
\newblock {Virtual to Real Reinforcement Learning for Autonomous Driving}.
\newblock In \emph{Proceedings of the British Machine Vision Conference
  (BMVC)}, 11.1--11.13. London, UK: {BMVA Press}.

\bibitem[{Peng et~al.(2017)Peng, Berseth, Yin, and Van De~Panne}]{8}
Peng, X.~B.; Berseth, G.; Yin, K.; and Van De~Panne, M. 2017.
\newblock {DeepLoco: Dynamic Locomotion Skills Using Hierarchical Deep
  Reinforcement Learning}.
\newblock \emph{ACM Transactions on Graphics}, 36(4): 1--13.

\bibitem[{Rasmussen and Williams(2006)}]{21}
Rasmussen, C.~E.; and Williams, C. K.~I., eds. 2006.
\newblock \emph{Gaussian Processes for Machine Learning}.
\newblock Cambridge, MA, USA: MIT Press.

\bibitem[{Richter and Roy(2017)}]{16}
Richter, C.; and Roy, N. 2017.
\newblock {Safe Visual Navigation via Deep Learning and Novelty Detection}.
\newblock In \emph{Proceedings of Robotics: Science and Systems}, 1--9.
  Cambridge, MA, USA: {MIT Press}.

\bibitem[{Silver et~al.(2016)Silver, Huang, Maddison, Guez, Sifre,
  Schrittwieser, Antonoglou, Panneershelvam, Lanctot, Dieleman, Grewe, Nham,
  Kalchbrenner, Sutskever, Lillicrap, Leach, Kavukcuoglu, Graepel, and
  Hassabis}]{4}
Silver, D.; Huang, A.; Maddison, C.~J.; Guez, A.; Sifre, G. v.~d.,
  Laurent~Driessche; Schrittwieser, J.; Antonoglou, I.; Panneershelvam, V.;
  Lanctot, M.; Dieleman, S.; Grewe, D.; Nham, J.; Kalchbrenner, N.; Sutskever,
  I.; Lillicrap, T.; Leach, M.; Kavukcuoglu, K.; Graepel, T.; and Hassabis, D.
  2016.
\newblock {Mastering the game of Go with deep neuralnetworks and tree search}.
\newblock \emph{Nature}, 529(7587): 484–489.

\bibitem[{Sofman et~al.(2011)Sofman, Neuman, Stentz, and Bagnell}]{34}
Sofman, B.; Neuman, B.; Stentz, A.; and Bagnell, J.~A. 2011.
\newblock {Anytime online novelty and change detection for mobile robots}.
\newblock \emph{Journal of Field Robotics}, 28(4): 589--618.

\bibitem[{Szegedy et~al.(2014)Szegedy, Zaremba, Sutskever, Bruna, Erhan,
  Goodfellow, and Fergus}]{18}
Szegedy, C.; Zaremba, W.; Sutskever, I.; Bruna, J.; Erhan, D.; Goodfellow, I.;
  and Fergus, R. 2014.
\newblock {Intriguing Properties of Neural Networks}.
\newblock In \emph{2nd International Conference on Learning Representations,
  ICLR 2014}, 1--10. La Jolla, CA, USA: {iclr.cc}.

\bibitem[{Tessler, Mankowitz, and Mannor(2018)}]{10}
Tessler, C.; Mankowitz, D.~J.; and Mannor, S. 2018.
\newblock Reward Constrained Policy Optimization.
\newblock arXiv:1805.11074.

\bibitem[{Thananjeyan et~al.(2021)Thananjeyan, Balakrishna, Nair, Luo,
  Srinivasan, Hwang, Gonzalez, Ibarz, Finn, and Goldberg}]{17}
Thananjeyan, B.; Balakrishna, A.; Nair, S.; Luo, M.; Srinivasan, K.; Hwang, M.;
  Gonzalez, J.~E.; Ibarz, J.; Finn, C.; and Goldberg, K. 2021.
\newblock {Recovery RL: Safe Reinforcement Learning With Learned Recovery
  Zones}.
\newblock \emph{IEEE Robotics and Automation Letters}, 6(3): 4915--4922.

\bibitem[{Thananjeyan et~al.(2020)Thananjeyan, Balakrishna, Rosolia, Li,
  McAllister, Gonzalez, Levine, Borrelli, and Goldberg}]{14}
Thananjeyan, B.; Balakrishna, A.; Rosolia, U.; Li, F.; McAllister, R.;
  Gonzalez, J.~E.; Levine, S.; Borrelli, F.; and Goldberg, K. 2020.
\newblock {Safety Augmented Value Estimation From Demonstrations (SAVED): Safe
  Deep Model-Based RL for Sparse Cost Robotic Tasks}.
\newblock \emph{IEEE Robotics and Automation Letters}, 5(2): 3612--3619.

\bibitem[{Zenke, Poole, and Ganguli(2017)}]{32}
Zenke, F.; Poole, B.; and Ganguli, S. 2017.
\newblock {Continual Learning Through Synaptic Intelligence}.
\newblock In \emph{Proceedings of the 34th International Conference on Machine
  Learning}, 3987--3995. Bletchley Park, UK: {PMLR}.

\end{thebibliography}

\end{document}

\maketitle

% ----- Section A
\section{Gaussian Process Regression}

An important component of the proposed data-driven, danger-mitigation system is to construct a model of the relationship, \( \textbf{y} = f(\textbf{x}) \), between the uncertainty measure, \( \textbf{y} \), being used to drive the optimization and the action space, \( \textbf{x} \). To accomplish this, Gaussian process (GP) regression (Rasmussen and Williams, 2006) is used. One of the benefits of GP regression is that a functional form for the regression model does not need to be specified. Instead, GP regression assumes that any finite subset of observations \( S_y=\{y_1,y_2,...,y_n \}=\{f(x_1 ),f(x_2 ),...,f(x_n )\} \) are generated from a Gaussian process and, thus, follow a multivariate Gaussian distribution.

A given set of $ n $ observations, $ S_y $, can therefore be interpreted as a single, $ n $-dimensional point sampled from an $ n $-variate Gaussian distribution, with mean function, \( \mu(\textbf{x}) \), and variance function, \( \sigma^2(\textbf{x}) \). A covariance function, \( k(x_a, x_b) \), associated with this probability distribution specifies how any pair of observations, \( \{y_a, y_b\} \), are related. This functional relationship, often referred to as a \textit{kernel function}, is a type of distance measure. Two observations that are close to each other in the domain of \( f(\textbf{x}) \) will register a higher covariance value than two points that are located further apart. They should therefore be expected to have a stronger correlation and to be similar in value as well. The kernel function is user-specified, and several popular choices exist in the literature (Rasmussen and Williams, 2006).

Given the covariance values between all pairs of observations, it is possible to compute an estimate of the observation at some new location, $ x^* $, based simply on the proximity of $ y^* $ with the available data, \( S_y \). In this way, GP regression allows the available data to ``speak for itself'' in terms of what it would expect the value $ y^* $ to be, given its location, $ x^* $, in relation to the locations, \( S_x={x_1,x_2,...,x_n }, \) of all the other data-points.

In a typical use-case scenario, a set of $ n $ observations, $ S_y $, with user-specified sampling noise, \( \sigma_{noise}^2 \), would be available, and a prediction of the value $ y^* $ at some new location $ x^* $ would need to be computed. Given the above, the value of $ y^* $ would therefore follow a probability distribution given by:

\begin{equation}
    P(y^* | \textbf{Py}, \textbf{Px}, x^*) = N(\mu(x^*), \sigma^2(x^*)),
    \label{eqA1}
\end{equation}

where:

\begin{gather}
    \mu(x^*) = \textbf{k}^*\textbf{K}^{-1}\textbf{Py}, \label{eqA2} \\
    \sigma^2(x^*) = k(x^*, x^*) - \textbf{k}^*\textbf{K}^{-1}\textbf{k}^{*\textrm{T}}, \label{eqA3}\\
    \textbf{K} = 
        \begin{bmatrix}
            k(x_1, x_1) & ... & k(x_1, x_n) \\
            \vdots & \ddots & \vdots \\
            k(x_n, x_1) & ... & k(x_n, x_n) \\
        \end{bmatrix}_{n \times n}
        + \sigma^2_{noise}I, \label{eqA4}\\
    \textbf{k}^* = \begin{bmatrix} k(x^*, x_1) & k(x^*, x_2) & ... & k(x^*, x_n) \end{bmatrix}_{1 \times n}, \label{eqA5}\\
    \textbf{Px} = \begin{bmatrix} x_1 & x_2 & ... & x_n \end{bmatrix}_{1 \times n}, \label{eqA6}\\
    \textbf{Py} = \begin{bmatrix} y_1 & y_2 & ... & y_n \end{bmatrix}_{1 \times n}, \label{eqA7}
\end{gather}

Here, $ N $ denotes the standard normal distribution and T denotes matrix-transpose. Derivations for equations \ref{eqA2} and \ref{eqA3} can be found in (Rasmussen and Williams, 2006). Upon solving the above, Eq. \ref{eqA2} serves as the best estimate for the observation $ y^* $, while Eq. \ref{eqA3} gives a measure of the uncertainty in this estimate. As new data becomes available and the set $ S_y $ grows, more information about the underlying relationship between \textbf{x} and \textbf{y} is made available, which updates and improves the model being constructed for the true relationship, \( \textbf{y} = f(\textbf{x}) \). This, in turn, improves the estimates of observations at unknown locations within the vicinity of the available data.

% ----- Section B
\section{The Variational Auto-Encoder}

\begin{figure}[ht]
    \centering
    \includegraphics[width=0.6\textwidth]{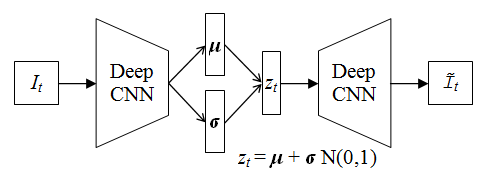}
    %\captionsetup{labelformat=empty}
    \caption{Structure of a Variational Auto-Encoder.}
    \label{figB1}
\end{figure}

In complex environments, observations obtained are typically high-dimensional (e.g., RGB images from camera sensors). In the proposed adaptation method presented in the main paper, a Variational Auto-Encoder (VAE) (Kingma and Welling, 2014) is used to both reduce the dimensionality of observation inputs and to provide a measure of observation uncertainty. A VAE, shown schematically in Figure \ref{figB1}, is composed of two main parts: an encoding stage and a decoding stage. The encoding stage consists of a deep convolutional neural network that reduces the high-dimensional observation input, \( I_t \), obtained at some time, $ t $, into a lower dimensional vector, $ z_t $, referred to herein as a \textit{latent vector}.

The decoder is another deep convolutional neural network that uses this latent vector to reconstruct the original input as best as possible. The VAE learns the parameters, \( \bm{\mu} \) and \( \bm{\sigma} \), of a multivariate Gaussian that represents the distribution of the latent space. Latent vectors are then sampled from this distribution during decoding to build the reconstruction. The VAE is trained in an unsupervised manner through stochastic gradient descent, using a dataset of observations taken from the environment, so as to minimize a measure of loss that includes the error between the output, \( \widetilde{I}_t \), of the decoder and the input, $ I_t $, to the encoder, as well as the divergence of the latent distribution parameters from a standard multivariate normal distribution.

\begin{figure}[ht]
    \centering
    \includegraphics[width=0.9\textwidth]{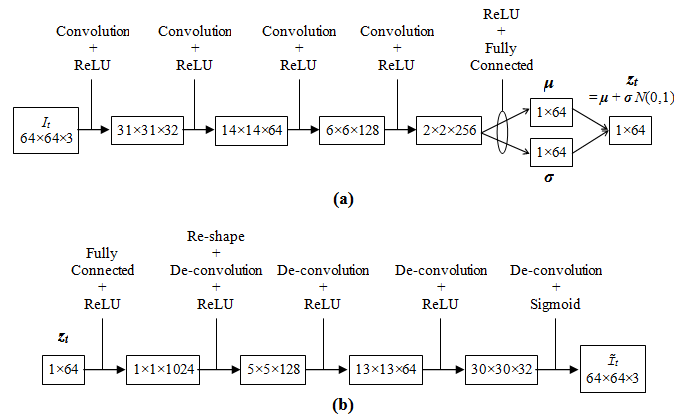}
    %\captionsetup{labelformat=empty}
    \caption{Neural network architecture of the Variational Auto-Encoder used in the experiments: (a) Encoder; (b) Decoder.}
    \label{figB2}
\end{figure}

The deep neural network structures of the encoder and decoder of the VAE constructed in Ha and Schmidhuber (2018) was used in the simulation experiments conducted for this paper as well. This design was chosen due to its success with encoding RGB color image observations from the OpenAI Gym game environment, Car Racing, and VizDoom. Figure \ref{figB2} shows the neural network architecture of this VAE. This model was trained using a total of 72000 RGB images obtained from 1 hour (simulation time) of driving using the CARLA auto-pilot within the urban environment map used for the simulations. A total of 182 epochs of training were used, with all 72000 images presented at each epoch in batches of 32.

% ----- Section C
\section{Establishing an Upper Bound on Reconstruction Error Based on Confidence Limits}

Another approach to establishing a value for the maximum threshold, $ ULe $, for the VAE reconstruction errors for emergency detection can also be to devise a probability distribution, \( P_n(e | \textbf{\textit{I}}) \), on the expected (nominal) VAE reconstruction error, $ e $, using probability distribution fitting techniques (Johnson, Kotz, and Balakrishnan, 1994) on the error data computed from the set of available relevant observation data, \( \textbf{\textit{I}} \). The user is free to select the particular type of distribution to use to best match the data. Once the reconstruction error probability distribution function has been constructed, it can be used to compute a \( \rho\% \), one-sided confidence limit, \( e_\rho \), on the expected reconstruction error, which will then represent the threshold, $ ULe $ (Eq. \ref{eqC1}).
\begin{equation}
    ULe = e_\rho : P_n(e < e_\rho | \textbf{\textit{I}}) = \rho
    \label{eqC1}
\end{equation}

\begin{figure}[ht]
    \centering
    \includegraphics[width=0.9\textwidth]{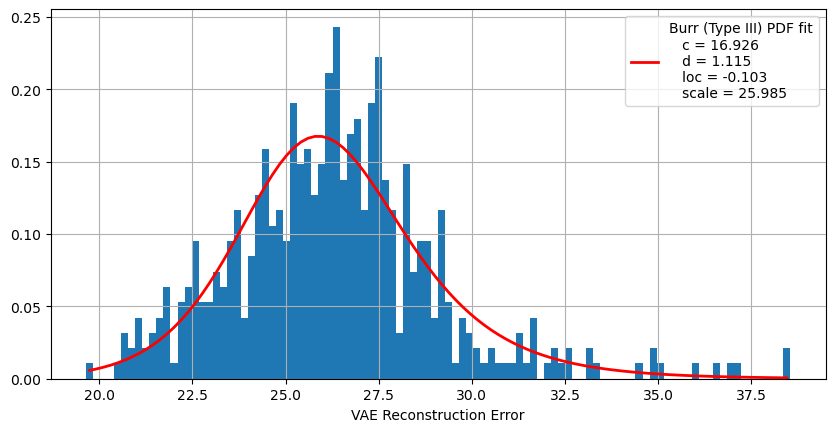}
    %\captionsetup{labelformat=empty}
    \caption{Example of a Burr (Type III) probability density function fitted to VAE reconstruction errors under nominal conditions.}
    \label{figC1}
\end{figure}

As an example, consider the autonomous driving car collision scenario 1 (see Figure 2a) used for the experiments that were presented in Section 4. In the absence of the stationary obstacle, and with the agent being controlled via the CARLA auto-pilot, Figure \ref{figC1} shows the distribution of VAE reconstruction errors (i.e., histogram of blue bars) obtained from observations made as the agent drives down the corresponding section of the road. The VAE used here is one that has already been trained to reconstruct observations accurately in this part of the environment map. However, given that the VAE learns the parameters of a distribution from which to sample the latent vectors for reconstruction, and that any given reconstruction will not be perfect, we can expect there to be some variation in the resulting errors under these nominal conditions. Figure \ref{figC1} shows a Burr (Type III) probability density function (Burr, 1942) fitted to raw VAE reconstruction error data (using the statistics package from the SciPy library for Python), computed as the mean squared error between the input RGB image, $ I_t $, and the reconstructed output, \( \widetilde{I}_t \). This probability density function for this particular fit is defined by two shape parameters, $c$ = 16.926 and $d$ = 1.115, and is generalized through the location parameter, $loc$ = -0.103, and the scale parameter, $ scale $ = 25.985.

\begin{figure}[ht]
    \centering
    \includegraphics[width=0.9\textwidth]{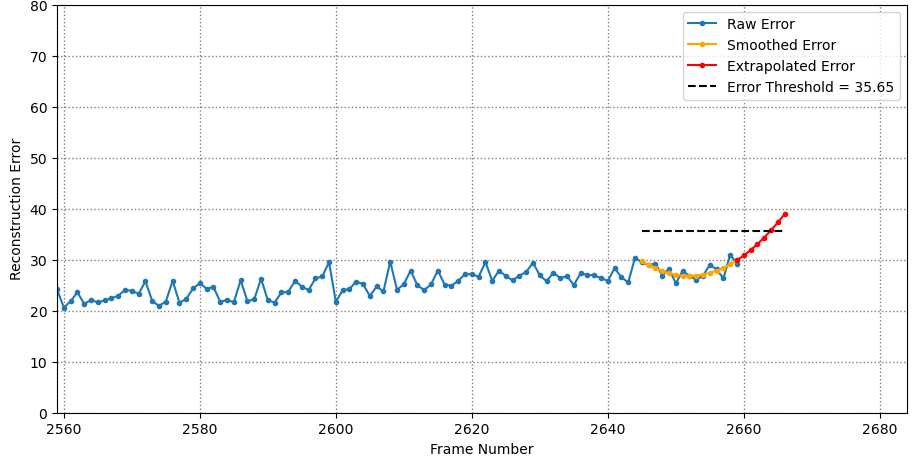}
    %\captionsetup{labelformat=empty}
    \caption{Example of a positive emergency detection based on a 99.5\% one-sided confidence limit, giving $ ULe $ = 35.65.}
    \label{figC2}
\end{figure}

We can subsequently compute a one-sided, \( \rho = 99.5\% \) confidence limit, \( e_{99.5\%} \), for example, which gives \( e_{99.5\%} = 35.65 \), where \( P_n(e < 35.65 | \textbf{\textit{I}}) = 0.995 \). Figure \ref{figC2} shows a plot of VAE reconstruction errors with respect to time-step during a sample simulation experiment within Scenario 1, at the point where a positive detection is made. This positive detection comes from a 7-time-step extrapolation (red curve) of a second-order polynomial regression fit, \( f_e(t) \), to the last 15 errors computed (orange curve), where the last $ q $=3 extrapolated errors exceed the established $ ULe $ = 35.65 threshold. For reference, the $ ULe $ threshold used in the Scenario 1 experiments for the paper was 35.

% ----- Section D
\section{Simulation Experiment Settings}

The simulation experiments were conducted in the CARLA (v0.9.6) autonomous driving car simulator environment. The autonomous agent was equipped with a forward-facing RGB camera that provided observations in the form of \( 64\times64\times3 \) images. The camera was mounted sufficiently forward such that no part of the car was visible in any of the observation images. The simulation was run at a speed of 20 frames per second. All experiments were performed on an Ubuntu 18.04 system equipped with an Nvidia GeForce RTX 2080 Ti GPU, a 3.60GHz Intel{\textregistered} Xeon{\textregistered} Gold 5122 CPU, and 63GB of memory. The primary programming language was Python (v3.7).

In order to consider the worst-case scenario, the default behavior of the agent upon encountering an obstacle was to take no action. To accomplish this, the simulated agent was controlled via an auto-pilot built into the CARLA simulator, but with all collision detection and avoidance functionality turned off. As such, collision avoidance was completely the responsibility of the emergency response action-sequence employed. At any given time-step, $ i $, the agent specifies the throttle, braking, and steering inputs through a 2D, real, action-vector, \( \textbf{a}_i = [a_{i_1}, a_{i_2}] \), where each value ranges from 0 to 1 and is interpreted as described in Table \ref{tableD1}.

\begin{table}[t]
    \fontsize{9}{12}\selectfont
    \centering
    \begin{tabular}{|m{0.09\columnwidth}<{\centering}|m{0.12\columnwidth}<{\centering}|m{0.13\columnwidth}<{\centering}|m{0.48\columnwidth}<{\centering}|}
        \hline
        \textbf{Element} & \textbf{Control} & \textbf{Value/Range} & \textbf{Interpretation} \\
        \hline
        \multirow{5}*{\( a_{i_1} \)} & \multirow{5}*{Acceleration} & 0 & Full braking; No throttle \\
        \cline{3-4}
         & & (0, 0.5) & Linearly decreasing braking intensity from full- to no-braking; No throttle \\
        \cline{3-4}
         & & 0.5 & No braking; No throttle \\
        \cline{3-4}
         & & (0.5, 1) & Linearly increasing throttle intensity from no- to full-throttle; No braking \\
        \cline{3-4}
         & & 1 & Full throttle; No braking \\
        
        \hline
        
        \multirow{5}*{\( a_{i_2} \)} & \multirow{5}*{Steering} & 0 & Full left steer \\
        \cline{3-4}
         & & (0, 0.5) & Linearly proportional steering angle from full-left to straight \\
        \cline{3-4}
         & & 0.5 & Wheels straight \\
        \cline{3-4}
         & & (0.5, 1) & Linearly proportional steering angle from straight to full-right \\
        \cline{3-4}
         & & 1 & Full right steer \\
        \hline
    \end{tabular}
    %\captionsetup{labelformat=empty}
    \caption{Interpretation of action-vector components.}
    \label{tableD1}
\end{table}

For emergency detection, the VAE used was trained using a dataset of images obtained from a pre-deployment simulation, wherein the agent was made to drive on the same CARLA urban environment map used in the simulation experiments, but without any obstacles. The agent was allowed to drive using the auto-pilot over the entire map (with all road choices being made randomly) for 1 hour in simulation time. This image dataset was used with the PyTorch (v1.4.0) machine learning library to train the VAE. This pre-training was done to represent existing knowledge that the agent possesses, before it is introduced to something very different and relative to which the unforeseen event is compared.

For response generation, the BoTorch (v0.2.1) Bayesian Optimization library, which is built on Python, PyTorch, and GPyTorch, was used to construct the GP regression model and to optimize the UCB acquisition function during each iteration of the BO loop. The Mat\'ern kernel function (Rasmussen and Williams, 2006) was used when solving the GP regression, with the fixed sampling noise, \( \sigma^2_{noise} \), taken to be 0.01. At the start of each response action-sequence, the mean function, \( \mu(\textbf{x}) \), of the GP model was set to zero everywhere, and the variance function, \( \sigma^2(\textbf{x}) \), was set to a constant 0.01.

Table \ref{tableD2} summarizes the number of simulation repetitions performed as well as the upper reconstruction error threshold, $ ULe $, used for each scenario tested at the average agent speed of 20km/h, where the proposed methods for emergency detection and response generation were applied. Table \ref{tableD2} also shows the number of simulation repetitions used for each scenario under the 30km/h tests, where the response generation procedure only was being tested (i.e., where the proposed method and random response generation were triggered manually at the same location relative to the obstacles for a fair comparison).

\begin{table}[t]
    \centering
    \begin{tabular}{|m{0.12\columnwidth}<{\centering}|m{0.12\columnwidth}<{\centering}|m{0.12\columnwidth}<{\centering}|m{0.22\columnwidth}<{\centering}|}
        \hline
        \multirow{2}*{\textbf{Scenario \#}} & \multicolumn{2}{|c|}{\textbf{20km/h Tests}} & \multirow{2}*{\begin{tabular}{@{}c@{}} \textbf{No. of Replications} \\ \textbf{for 30 km/h Tests} \end{tabular}} \\
        \cline{2-3}
         & \textbf{No. of Replications} & \textbf{\textit{ULe}} & \\
        \hline
        1 & 20 & 35 & 100 \\
        \hline
        2 & 20 & 38 & 20 \\
        \hline
        3 & 20 & 38 & 20 \\
        \hline
        4 & 20 & 38 & 20 \\
        \hline
        5 & 20 & 38 & 20 \\
        \hline
        6 & 20 & 29 & 20 \\
        \hline
        7 & 20 & 29 & 20 \\
        \hline
        8 & 20 & 32 & 40 \\
        \hline
        9 & 20 & 32 & 40 \\
        \hline
    \end{tabular}
    \caption{Replication and threshold settings for 20km/h and 30km/h tests.}
    \label{tableD2}
\end{table}